\documentclass{article}

\usepackage[utf8]{inputenc}
\usepackage{caption}
\usepackage{graphicx}
\usepackage{wrapfig}
\usepackage{subcaption}
\usepackage{amsmath}
\usepackage{amssymb}
\usepackage{hyperref}
\usepackage{csquotes}
\usepackage{booktabs}
\usepackage[margin=1.25in]{geometry}

\newcommand\blfootnote[1]{%
  \begingroup
  \renewcommand\thefootnote{}\footnote{#1}%
  \addtocounter{footnote}{-1}%
  \endgroup
}

\title{Reasoning and Generalization in RL:\\A Tool Use Perspective}
\author{
 Sam Wenke \\
 Fomoro AI \\
 \small{\texttt{sam@fomoro.com}} \\
 \and
 Dan Saunders \\
 Fomoro AI \\
 \small{\texttt{dan@fomoro.com}} \\
 \and
 Mike Qiu \\
 Fomoro AI \\
 \small{\texttt{mike@fomoro.com}} \\
 \and
 Jim Fleming \\
 Fomoro AI \\
 \small{\texttt{jim@fomoro.com}} \\
}

\begin{document}
\maketitle

\begin{abstract}
\noindent{Learning to use tools to solve a variety of tasks is an innate ability of humans and has been observed of animals in the wild. However, the underlying mechanisms that are required to learn to use tools are abstract and widely contested in the literature. In this paper, we study tool use in the context of reinforcement learning and propose a framework for analyzing generalization inspired by a classic study of tool using behavior, the trap-tube task. Recently, it has become common in reinforcement learning to measure generalization performance on a single test set of environments. We instead propose transfers that produce multiple test sets that are used to measure specified types of generalization, inspired by abilities demonstrated by animal and human tool users. The source code to reproduce our experiments is publicly available at \url{https://github.com/fomorians/gym_tool_use}.}
\end{abstract}

\blfootnote{All authors helped to write and edit the paper. Sam conceived the role of tool use experiments in reinforcement learning, and with Mike's help, conducted a thorough literature review, which was used to write large portions of the paper. Sam devised the experimental setup, and Dan ran the experiments and collated the results. Jim wrote the model and algorithm code, and with Sam's help, set the overall direction of the research.}

\section{Introduction}

Tool use is considered a hallmark of animal and human intelligence. Analysis of tool-using behaviors have unveiled useful information about the cognitive mechanisms of various species. For example, woodpecker finches on the Galapogos islands are able to use tools through learning by trial and error \cite{Tebbich2010}, while wild bottlenose dolphins appear to use marine sponges as foraging tools \cite{Krutzen2005}. Furthermore, wild gorillas have been observed using branches to test the depth of water, as an aid in walking in deep water, and a bridge to cross a patch of swamp \cite{Breuer2005}. The sophisticated use of tools demand potentially complex cognitive reasoning abilities \cite{Teschke2013}.

\textit{Causal}, \textit{relational} (\textit{transitive}), \textit{analogical}, and \textit{symbolic} reasoning have all been implicated in the cognitive processes of tool users. Causal reasoning has been linked to the cognitive abilities of tool using animals and humans \cite{Visalberghi1994, McCormack2011}, though some argue that only humans rely on ``higher-order'' causal relations to make sense of the world \cite{Reaux2003, Povinelli2011}. They suggest that humans possess a uniquely systematic, \textit{domain-general} relational reasoning ability, whereas other animal tool users rely on \textit{domain-specific} expectations for problem solving \cite{Penn2008}. Symbolic reasoning requires the subject to represent a symbol (such as a drawing) as an actual object and as something else that the symbol stands for (\textit{dual representation} \cite{DeLoache1997}). This quality is naturally acquired by children at a young age, and may contribute to the generalization of a solution to a wide range of new contexts\cite{Seed2010}. To be considered a tool user, an animal (or agent) must demonstrate their ability to systematically solve a variety of reasoning-specific and domain-general tasks.

Learning to use tools to solve a variety of tasks is an innate ability of humans and has been speculatively observed of animals in the wild. The underlying mechanisms that are required to learn to use tools are abstract and widely contested in literature. Models of the requirements of tool usage inspired by animals and humans have been proposed, such as grasping or manipulation in robots \cite{Li1988, Shimoga1996, Yamashita1998, Gupta1998, Halperin2000}, combining hard-coded exploratory behaviors to perform more dynamic skills \cite{Stoytchev2005}, and Inductive Logic Programming \cite{Brown2012, Wicaksono2016, Wicaksono2017}. More complex models learn to plan future trajectories to perform saw and cut behaviors \cite{Lenz2015}, grasp and manipulate tools for new tasks \cite{Fang2018}, and utilize previously unseen objects from demonstrations and self-supervision through model-free \cite{Rajeswaran2017} and model-based control \cite{Ebert2017, Ebert2018, Xie2019}. Tool use generalization is still broadly unsolved and prior works lack the framework to properly measure it. In this paper, we propose a benchmark for analyzing tool use and assessing generalization in a discrete environment inspired by a classic experiment of animal tool using behavior.

We introduce a framework for evaluating the generalization of methods that learn to use tools in terms of the reinforcement learning (RL) framework \cite{Sutton2018}. Our contributions are inspired by prior work studying the generalization of deep RL methods due to their notable ability to learn complex, highly successful control algorithms. The generalization capabilities of RL algorithms is interesting due to the lack of supervision given to agents in their training environments and the expectation that the agent generalizes to possibly unseen environments and tasks. Learning policies that are sensitive to changes in the environment and can adapt to these variations is an important approach to improving the generalization ability of RL algorithms. For example, the introduction of different modes of emulated Atari 2600 games allowed for agents to be trained in one environment while being evaluated in slightly ones \cite{Farebrother2018}. Systematically sampling the parameters of the dynamics of an environment measures the ability of RL algorithms to generalize to test environments that are similar to or different from the training environments \cite{Packer2018}. Most similar to our work, attempts at quantifying generalization have been made by carefully designing a set of procedurally generated environments that are split into training and testing sets \cite{Cobbe2018}.

Our generalization study, inspired by tool use experiments and the complex reasoning abilities of animals, categorizes generalization into multiple test environments that can be used to evaluate a particular set of reasoning abilities of learned agents. To introduce our framework, we define a set of simply represented discrete environments that test the use of a tool to solve domain-general variations of an interpretation of a classic animal tool use study.

\subsection{Tool Use}

For the purpose of our study, we present our working definition of tool use and how it relates to our study of RL. The definition of tool use has evolved over time to account for new observations of animal tool use as well as to better approximate our intuitive understanding on what it means to use tools. We use the definition from \cite{StAmant2008} to frame our study, which can be found in \nameref{section:appendicies_definitions}. The manipulation of an object by the agent and the completion of a task in an environment is an important objective of RL and is the basis of our study.

\section{Methods}

We adapt a classic experiment from the study of tool use behavior, called the \textit{trap-tube task} \cite{Visalberghi1994}. Subjects of the trap-tube task are presented with a transparent tube. Within the tube there is a trap in the center and an object (representing a reward) placed next to the trap and out-of-reach of the subject. A stick of a fixed length and a diameter smaller than the tube is placed next to the subject. The stick represents a tool that can be used to push the object away from the trap and within reach of the subject from outside of the tube.

Variants of the trap-tube experiment have helped tease apart the cognitive mechanisms involved in complex tool use. Studies have shown that most chimpanzees, capuchin monkeys and New Caledonian crows rely on \textit{perceptual knowledge} to solve the trap-tube task \cite{Visalberghi1994, Kacelnik2009b}. That is, subjects that learned to solve the base task continued to avoid the non-functional trap when the tube was inverted, suggesting that their solutions were informed by the perceptual features of the trap rather than the understanding of its physical properties. However, a small subset of chimpanzees and crows demonstrated physical reasoning abilities by solving a modified trap task where the sides of the tube are blocked and the food can only be pushed down a hole to an exit beneath the tube \cite{Seed2009}. In this setup, subjects require \textit{structural knowledge} to solve the task because they need to understand the functional significance of traps and barriers. Finally, a covered version of the trap task, in which stickers were placed in the same location as the traps and barriers, reveals the subjects’ ability to use \textit{symbolic knowledge}. Here, subjects have to reason symbolically by interpreting the stickers as their respective structures \cite{Seed2011}. 

According to Seed et al., \cite{Seed2011} the use of \textit{perceptual}, \textit{structural}, and \textit{symbolic} knowledge requires progressively deeper levels of abstraction. We stick to this proposition and individually redefine these requirements as categories of generalization across a variety of reasoning-specific and domain-general tasks. We use the definitions for the requirements of generalization Seed et al. \cite{Seed2011}:

\begin{itemize}
\item \textbf{Perceptual}: Generalization across stimuli that share \textit{perceptual features}.
\item \textbf{Structural}: Generalization across stimuli that share \textit{abstract, structural features}.
\item \textbf{Symbolic}: Generalization across stimuli that share \textit{abstract, conceptual features}.
\end{itemize}

An agent generalizes to a set of alterations to a ground truth task if it is able to succeed in both the ground truth and altered tasks. In order to test this interpretation, we introduce a set of simulated environments that are used to evaluate generalization to alterations of the classic trap-tube task.

\section{Environments}

We evaluate the generalization of the behavior of agents in environments given as RL problems known as \textit{Markov decision processes} (MDP, \cite{Sutton2018}). An MDP consists of a set of states $\mathcal{S}$, a set of actions $\mathcal{A}$, a transition kernel $\mathcal{T} : \mathcal{S}\times{}\mathcal{A}\rightarrow{}\triangle{}\mathcal{S}$, a reward function $\mathcal{R} : \mathcal{S}\times{}\mathcal{A}\rightarrow{}\mathbb{R}$, and an initial state $s_{0} \in{} \mathcal{S}$ drawn from a distribution $\mathcal{P} \in{} \triangle{}\mathcal{S}$. An agent interacts with the MDP sequentially: at each timestep it observes the current state $s \in{} \mathcal{S}$, takes an action $a \in{} \mathcal{A}$, transitions to the next state $s'$ drawn from the distribution $\mathcal{T}(s,\,a)$, and receives a reward $\mathcal{R}(s,\,a)$. Additionally, we propose a \textit{state transfer kernel} $\mathcal{F} : \mathcal{S} \rightarrow{} \mathcal{S'}$ to map a set of states of the MDP to another set of states.

\begin{figure}[t!]
    \centering
    \addtocounter{subfigure}{-1}
    \begin{subfigure}[t]{0.09\textwidth}
        \centering
        \captionsetup{labelformat=empty, skip=3pt}
        \includegraphics[width=.25\linewidth]{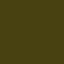}
        \caption{\texttt{Ground}}
    \end{subfigure}
    \addtocounter{subfigure}{-1}
    \begin{subfigure}[t]{0.09\textwidth}
        \centering
        \captionsetup{labelformat=empty, skip=3pt}
        \includegraphics[width=.25\linewidth]{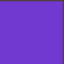}
        \caption{\texttt{Agent}}
    \end{subfigure}
    \begin{subfigure}[t]{0.09\textwidth}
        \centering
        \captionsetup{labelformat=empty, skip=3pt}
        \includegraphics[width=.25\linewidth]{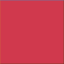}
        \caption{\texttt{Food}}
    \end{subfigure}
    \addtocounter{subfigure}{-1}
    \begin{subfigure}[t]{0.09\textwidth}
        \centering
        \captionsetup{labelformat=empty, skip=3pt}
        \includegraphics[width=.25\linewidth]{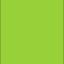}
        \caption{\texttt{Tool}}
    \end{subfigure}
    \addtocounter{subfigure}{-1}
    \begin{subfigure}[t]{0.18\textwidth}
        \centering
        \captionsetup{labelformat=empty, skip=3pt, justification=centering}
        \includegraphics[width=.13\linewidth]{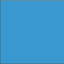}
        \caption{\texttt{Tube (Top) \\ Tube (Bottom)}}
    \end{subfigure}
    \addtocounter{subfigure}{-1}
    \begin{subfigure}[t]{0.09\textwidth}
        \centering
        \captionsetup{labelformat=empty, skip=3pt}
        \includegraphics[width=.25\linewidth]{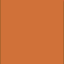}
        \caption{\texttt{Trap}}
    \end{subfigure}
    \addtocounter{subfigure}{-1}
    \begin{subfigure}[t]{0.09\textwidth}
        \centering
        \captionsetup{labelformat=empty, skip=3pt}
        \includegraphics[width=.25\linewidth]{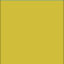}
        \caption{\texttt{Exit}}
    \end{subfigure}
    \addtocounter{subfigure}{-1}
    \addtocounter{subfigure}{-1}

    ~
    \\
    \vspace{5pt}

    \centering
    \begin{subfigure}[t]{0.29\textwidth}
        \captionsetup{width=.95\linewidth, singlelinecheck=false}
        \includegraphics[width=\textwidth]{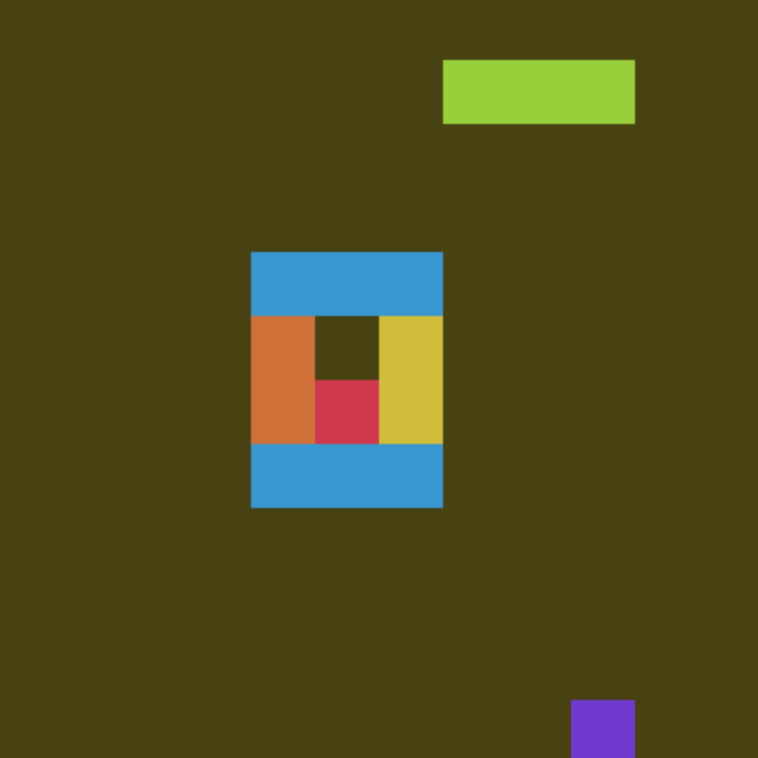}
        \caption{Initial state of the perceptual trap-tube environment. Object positions and rotations are randomized and regenerated each episode.}
    \end{subfigure}
    ~
    \begin{subfigure}[t]{0.29\textwidth}
        \captionsetup{width=.95\linewidth, singlelinecheck=false}
        \includegraphics[width=\textwidth]{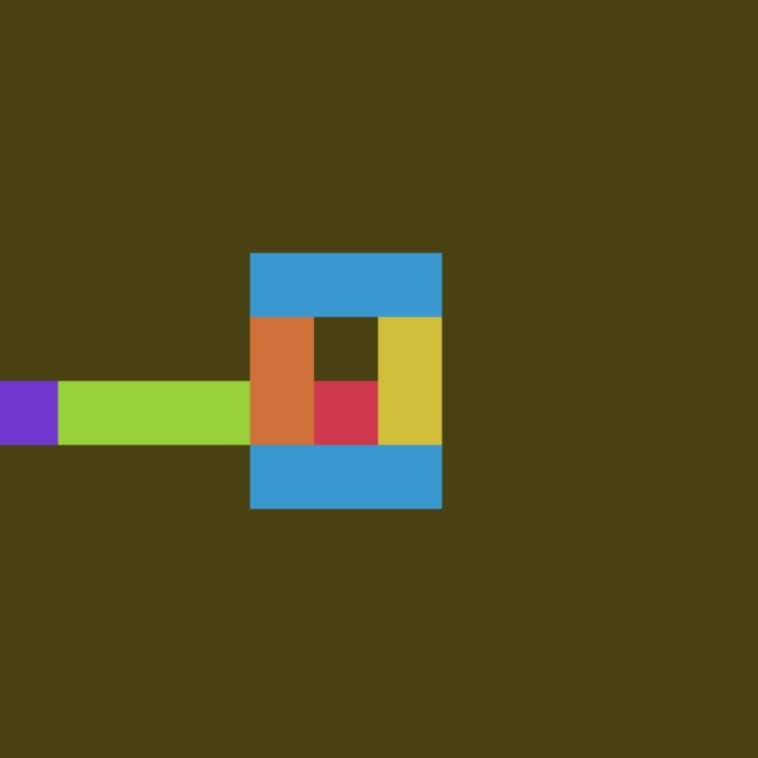}
        \caption{The agent locates the tool in the top right of the grid, and moves it to the left of the food and exit.}
    \end{subfigure}
    ~
    \begin{subfigure}[t]{0.29\textwidth}
        \captionsetup{width=.95\linewidth, singlelinecheck=false}
        \includegraphics[width=\textwidth]{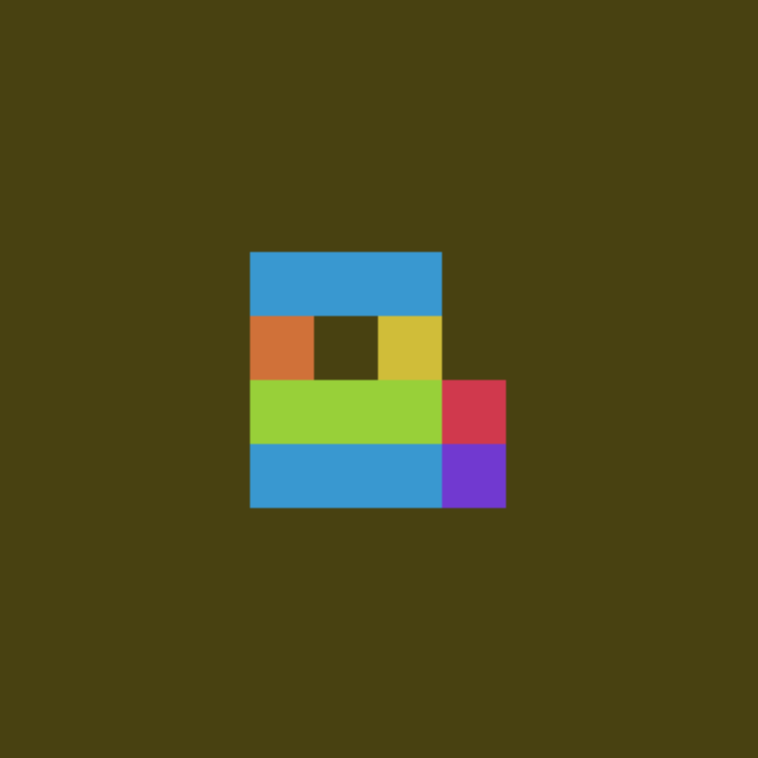}
        \caption{The agent uses the tool to move the food through the exit, out of the tube and within reach of the agent.}
    \end{subfigure}
    \caption{States of a perceptual trap-tube environment.}
    \label{fig:perceptual}
\end{figure}

Our focus is on clarifying environments represented in a simple form, as \textit{gridworlds}: a grid of cells, each occupied by an object with properties. Object properties are represented by a $k$-tuple: $(o_{0},\,o_{1},\,\ldots{},\,o_{k}) \in{} \mathbb{R}$. For simplicity, we choose $k = 3$ and can present the grid as an RGB image: $(o_{0} = R,\,o_{1} = G,\,o_{2} = B)$. All environments use a grid of size $12\times{}12$. At each time-step, the agent observes the entire grid as the current state $s$ of size $12\times{}12\times{}3$. An agent occupies one of the cells at any given time and can move and interact with adjacent objects as defined by the set of grasp actions $A_{G} = \{\leftarrow{}_g,\,\rightarrow{}_{g},\,\uparrow{}_{g},\,\downarrow{}_{g}\}$ and move actions $A_{M} =  \{\leftarrow{}_{m},\,\rightarrow{}_{m},\,\uparrow{}_{m},\,\downarrow{}_{m}\}$ that make up the action set of the environment $\mathcal{A} = A_{G} \times{} A_{M}$. Each move action changes the agent's position to the adjacent cell in the corresponding direction, and grasp actions allow the agent to move with the tool object given that the agent is adjacent to the tool and the grasp action is in the direction of the tool.

\subsection{Trap-Tube Environments} \label{section:trap_tube_environments}

The \textit{trap-tube} environments test an agent's ability to control a tool object with the goal of clearing the agent's path to a reward object. The reward object is surrounded by two tube barrier objects positioned opposite of one another, a trap barrier object and an exit object lie directly opposite of one another and adjacent to the tube barrier objects. The tool is initially located in a random position within the environment, always in reach of the agent. An agent will be able to move with the tool in a given direction, if it is adjacent to the tool and if a grasp action is made in the direction of the tool. The agent is not able to move or pass through a tube, exit, tool, or trap object, while the reward object is able to pass through the exit object. If the agent reaches the reward object, the agent is rewarded ($r = 1$) and the task is successfully completed. The agent is allowed to observe 50 states in sequence of the environment task, ending on failure to complete the task if the reward object is not reached ($r = 0$).

To solve a task that requires the use of a tool is, in general, difficult. The trap-tube environments are customized to test the generalization capability of agents. Alterations are made to the environment using the previously defined categories of generalization: perceptual ($P$), structural ($St$), and symbolic ($Sy$). For the experiments outlined in this paper, we choose a base set of states $\mathcal{S}_{base}$ such that  $P(\mathcal{S}_{base}\sim{}\triangle{}\mathcal{F}(\mathcal{S})) > 0$. This means that $\mathcal{S}_{base}$ has a non-zero probability of being sampled from any given transfer, and is used as the set of states for the base MDP.

\subsubsection{Perceptual}

The ability to perceive an object as a tool, regardless of the shape, position and rotation, along with surrounding objects, is an innate ability of humans and observed of wild animals. The \textit{perceptual transfer} $\mathcal{F}_{P}$ samples the position and orientation of the tool, the width, height, and orientation of the tube, the position of the trap and exit of the tube, the position of the agent, and the position of the food in the tube. An episode of a perceptual trap-tube environment is shown in Figure \ref{fig:perceptual}.

\subsubsection{Structural}

\begin{figure}[t!]
    \addtocounter{subfigure}{-1}
    \centering
    \begin{subfigure}[t]{0.09\textwidth}
        \centering
        \captionsetup{labelformat=empty, skip=3pt}
        \includegraphics[width=.25\linewidth]{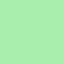}
        \caption{\texttt{Ground}}
    \end{subfigure}
    \addtocounter{subfigure}{-1}
    \begin{subfigure}[t]{0.09\textwidth}
        \centering
        \captionsetup{labelformat=empty, skip=3pt}
        \includegraphics[width=.25\linewidth]{img/agent.png}
        \caption{\texttt{Agent}}
    \end{subfigure}
    \begin{subfigure}[t]{0.09\textwidth}
        \centering
        \captionsetup{labelformat=empty, skip=3pt}
        \includegraphics[width=.25\linewidth]{img/food.png}
        \caption{\texttt{Food}}
    \end{subfigure}
    \addtocounter{subfigure}{-1}
    \begin{subfigure}[t]{0.09\textwidth}
        \centering
        \captionsetup{labelformat=empty, skip=3pt}
        \includegraphics[width=.25\linewidth]{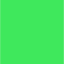}
        \caption{\texttt{Tool}}
    \end{subfigure}
    \addtocounter{subfigure}{-1}
    \begin{subfigure}[t]{0.18\textwidth}
        \centering
        \captionsetup{labelformat=empty, skip=3pt, justification=centering}
        \includegraphics[width=.13\linewidth]{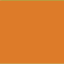}
        \caption{\texttt{Tube (Top) \\ Tube (Bottom)}}
    \end{subfigure}
    \addtocounter{subfigure}{-1}
    \begin{subfigure}[t]{0.09\textwidth}
        \centering
        \captionsetup{labelformat=empty, skip=3pt}
        \includegraphics[width=.25\linewidth]{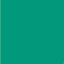}
        \caption{\texttt{Trap}}
    \end{subfigure}
    \addtocounter{subfigure}{-1}
    \begin{subfigure}[t]{0.09\textwidth}
        \centering
        \captionsetup{labelformat=empty, skip=3pt}
        \includegraphics[width=.25\linewidth]{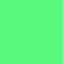}
        \caption{\texttt{Exit}}
    \end{subfigure}
    \addtocounter{subfigure}{-1}
    \addtocounter{subfigure}{-1}

    ~
    \\
    \vspace{5pt}

    \centering
    \begin{subfigure}[t]{0.29\textwidth}
        \captionsetup{width=.95\linewidth, singlelinecheck=false}
        \includegraphics[width=\textwidth]{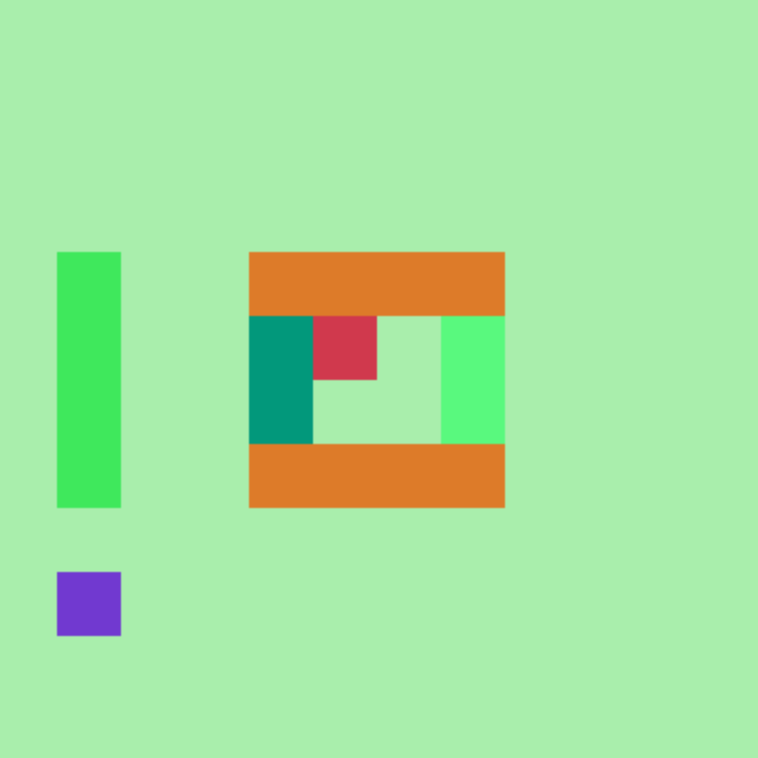}
        \caption{Initial state of the structural trap-tube environment. Object properties are randomized and regenerated each episode.}
    \end{subfigure}
    ~
    \begin{subfigure}[t]{0.29\textwidth}
        \captionsetup{width=.95\linewidth, singlelinecheck=false}
        \includegraphics[width=\textwidth]{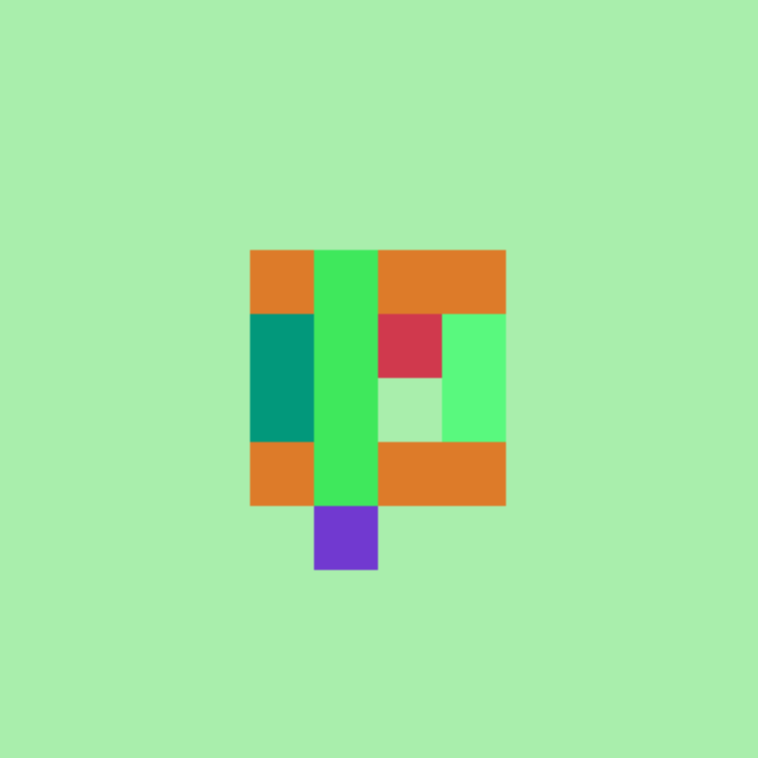}
        \caption{The agent locates, successfully grasps, and moves the tool towards the food and the exit.}
    \end{subfigure}
    ~
    \begin{subfigure}[t]{0.29\textwidth}
        \captionsetup{width=.95\linewidth, singlelinecheck=false}
        \includegraphics[width=\textwidth]{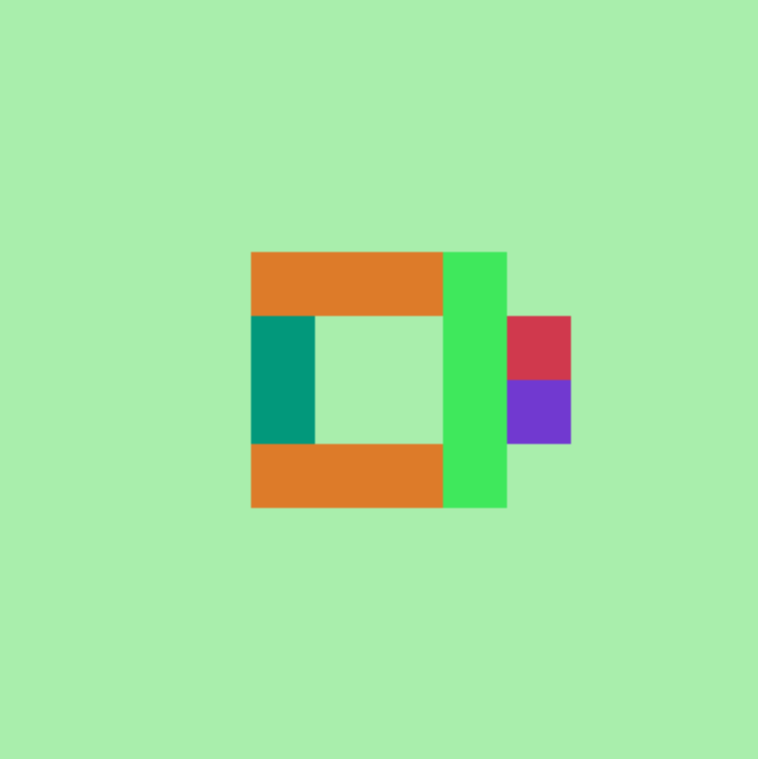}
        \caption{The agent uses the tool to move the food through the exit, out of the tube and within reach of the agent.}
    \end{subfigure}
    \caption{States of a structural trap-tube environment.}
    \label{fig:structural}
\end{figure}

Sensory signals interpreted by humans and animals are represented by multiple modalities, such as visual and auditory. The generalization of sensory feature representations is important to solve tool use tasks; for instance, an agent placed in a dark, color-less room is expected to solve a task that was originally demonstrated in a well-lit, colorful room. The \textit{structural transfer} $\mathcal{F}_{St}$ samples properties of the objects by category, excluding the agent and food (tool, trap, tube, exit, and ground). We represent the features of the environment as points on the surface of an $N$-dimensional spherical manifold. This was chosen to have an internal consistency in the representations to enable interpolation and to better represent how the real world is structured and represented consistently. $N$ is chosen to be 3 to form RGB colors for the purposes of rendering, thus the manifold is a sphere, however other values of $N$ should also work. An episode of a structural trap-tube environment is shown in Figure \ref{fig:structural}.

\subsubsection{Symbolic}

\begin{figure}[t!]
    \centering
    \begin{subfigure}[t]{0.09\textwidth}
        \centering
        \captionsetup{labelformat=empty, skip=3pt}
        \includegraphics[width=.25\linewidth]{img/ground.png}
        \caption{\texttt{Ground}}
    \end{subfigure}
    \addtocounter{subfigure}{-1}
    \begin{subfigure}[t]{0.09\textwidth}
        \centering
        \captionsetup{labelformat=empty, skip=3pt}
        \includegraphics[width=.25\linewidth]{img/agent.png}
        \caption{\texttt{Agent}}
    \end{subfigure}
    \begin{subfigure}[t]{0.09\textwidth}
        \centering
        \captionsetup{labelformat=empty, skip=3pt}
        \includegraphics[width=.25\linewidth]{img/food.png}
        \caption{\texttt{Food}}
    \end{subfigure}
    \addtocounter{subfigure}{-1}
    \begin{subfigure}[t]{0.16\textwidth}
        \centering
        \captionsetup{labelformat=empty, skip=3pt, justification=centering}
        \includegraphics[width=.14\linewidth]{img/tool.png}
        \caption{\texttt{Tube (Bottom)}}
    \end{subfigure}
    \addtocounter{subfigure}{-1}
    \begin{subfigure}[t]{0.13\textwidth}
        \centering
        \captionsetup{labelformat=empty, skip=3pt, justification=centering}
        \includegraphics[width=.17\linewidth]{img/tube.png}
        \caption{\texttt{Tube (Top) \\ Tool}}
    \end{subfigure}
    \addtocounter{subfigure}{-1}
    \begin{subfigure}[t]{0.09\textwidth}
        \centering
        \captionsetup{labelformat=empty, skip=3pt}
        \includegraphics[width=.25\linewidth]{img/trap.png}
        \caption{\texttt{Trap}}
    \end{subfigure}
    \addtocounter{subfigure}{-1}
    \begin{subfigure}[t]{0.09\textwidth}
        \centering
        \captionsetup{labelformat=empty, skip=3pt}
        \includegraphics[width=.25\linewidth]{img/exit.png}
        \caption{\texttt{Exit}}
    \end{subfigure}
    \addtocounter{subfigure}{-1}
    \addtocounter{subfigure}{-1}

    ~
    \\
    \vspace{5pt}

    \centering
    \begin{subfigure}[t]{0.29\textwidth}
        \captionsetup{width=.95\linewidth, singlelinecheck=false}
        \includegraphics[width=\textwidth]{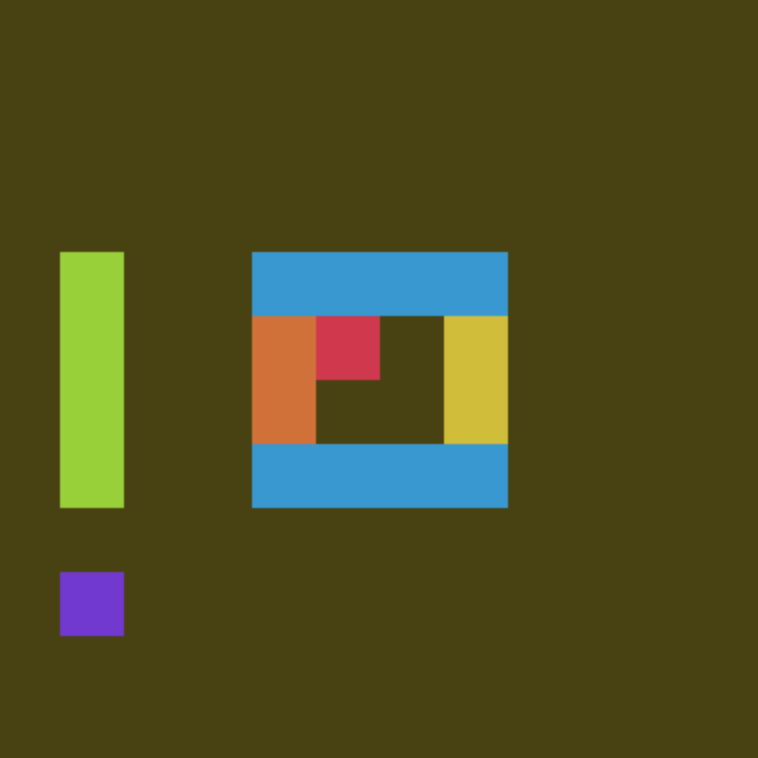}
        \caption{Initial state of the symbolic trap-tube environment. The base tool object is randomly swapped with another object, taking on the respective perceptual and structural properties of the objects.}
    \end{subfigure}
    ~
    \begin{subfigure}[t]{0.29\textwidth}
        \captionsetup{width=.95\linewidth, singlelinecheck=false}
        \includegraphics[width=\textwidth]{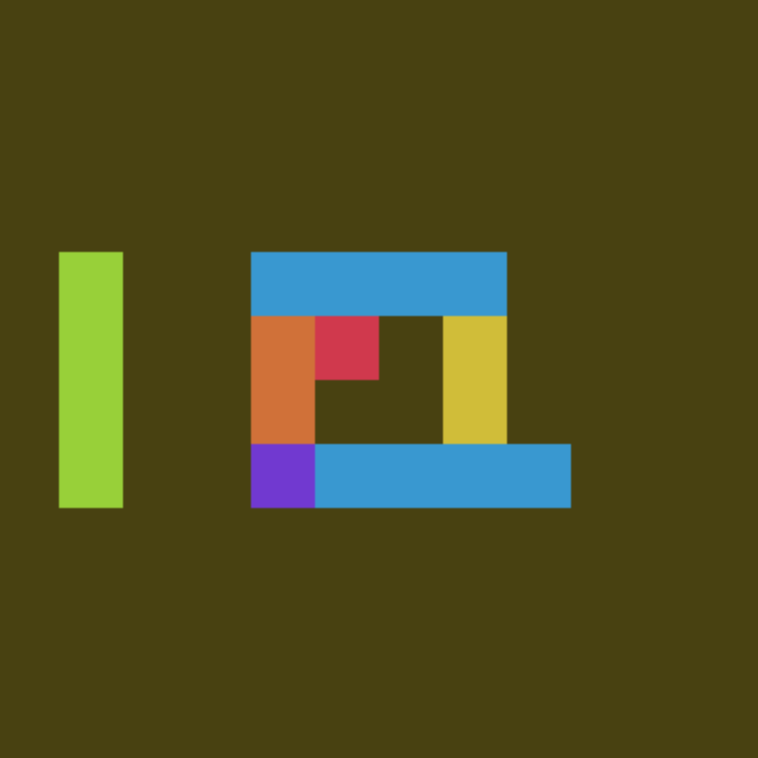}
        \caption{The agent locates the tool represented perceptually and structurally as the bottom of the tube. The agent proceeds to move the tool away from the direct path to the food.}
    \end{subfigure}
    ~
    \begin{subfigure}[t]{0.29\textwidth}
        \captionsetup{width=.95\linewidth, singlelinecheck=false}
        \includegraphics[width=\textwidth]{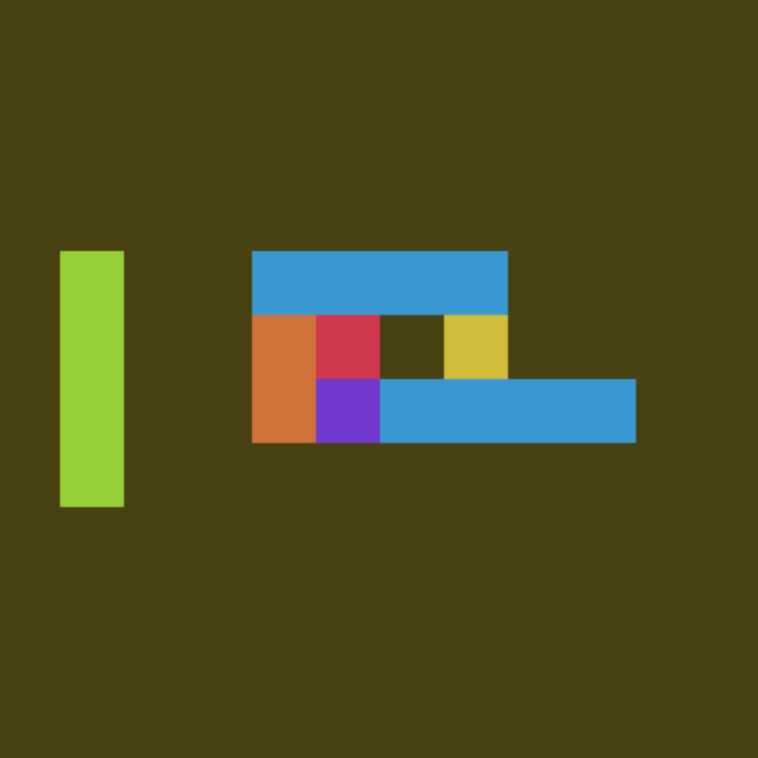}
        \caption{The agent moves towards the food. Although the tool is represented perceptually and structurally as the bottom of the tube, it still maintains the interactions expected of the tool, such as passing over the exit object.}
    \end{subfigure}
    \caption{States of a symbolic trap-tube environment.}
    \label{fig:symbolic}
\end{figure}

The interpretation that an object can arbitrarily be represented as a tool is an important innovation and intuition of humans. The generalization of the relationship between the agent, tool, obstacle, reward objects is important to understanding symbolic reasoning. In the trap-tube task, the \textit{symbolic transfer} $\mathcal{F}_{Sy}$ swaps the structural and perceptual properties of the base tool object with another object sampled from the set of possible objects: $\{\texttt{tool},\,\texttt{trap},\,\texttt{tube},\,\texttt{exit}\}$. A tool will behave the same, regardless of how it is represented structurally and perceptually, acting as a "symbol" of the environment to the agent. An episode of a symbolic trap-tube environment is shown in Figure \ref{fig:symbolic}.

\section{Experimental Setup}
\label{sec:exp_setup}

In order to measure the generalization of reinforcement learning algorithms for tool use, agents must be evaluated on all possible transfer sets $\mathcal{F}$: ($\{\mathcal{F}_{P}\}$, $\{\mathcal{F}_{St}\}$, $\{\mathcal{F}_{Sy}\}$, $\{\mathcal{F}_{P}, \mathcal{F}_{St}\}$, $\{\mathcal{F}_{P}, \mathcal{F}_{Sy}\}$, $\{\mathcal{F}_{St},\,\mathcal{F}_{Sy}\}$,$\{\mathcal{F}_{P},\,\mathcal{F}_{St},\,\mathcal{F}_{Sy}\}$). We outline the following experiments: (1) \textit{train agents on the base environment} and (2) \textit{train agents on the transfer set} $\{\mathcal{F}_{P},\,\mathcal{F}_{St},\,\mathcal{F}_{Sy}\}$. The goal of each experiment is to quantify the generalization of learned agents by measuring their success rate on all possible transfer sets. Experiment (1) challenges learned agents conditioned on a single task sampled from transfer set $\{\mathcal{F}_{P},\,\mathcal{F}_{St},\,\mathcal{F}_{Sy}\}$. Experiment (2) challenges learned agents to extrapolate reasoning abilities. Each experiment stresses the priors of learned agents. Successful agents must possess priors that enable learning to effectively reason perceptually, structurally, and symbolically.

\section{Results}

We present baseline results with a proximal policy optimization \cite{Schulman2017} (PPO) agent to use as a point of comparison for future work. Due to the sparsity of the task, we also include an intrinsic curiosity module \cite{Pathak2017} (ICM) to improve exploration. The agent observes the entire environment state and processes it into an embedding using two $2 \times 2$ convolutions, followed by two residual blocks and global spatial pooling to form an observation embedding. The agent's move and grasp actions from the previous time step are embedded into 8-dimensional vectors, while the reward from the previous time step is fed through an 8-unit dense layer to form a reward embedding. These embeddings are concatenated and fed to a 64-unit dense layer before finally feeding into a GRU \cite{Cho2014}. The GRU output is then used to compute action logits for move and grasp actions, a value prediction, and forward and inverse model predictions. The value prediction is used as the returns baseline while the forward and inverse model predictions are used for the ICM auxiliary losses and rewards. The agent is capable of discovering sparse rewards and learning to use the tool in the base tool use task, however, it fails to solve the generalization tasks. On both tasks, we train agents on 5 separate trials with a different random seed on each.

\subsection{Experiment 1}

\begin{figure}[t]
    \centering
    \begin{subfigure}[t]{0.45\textwidth}
        \includegraphics[width=\linewidth]{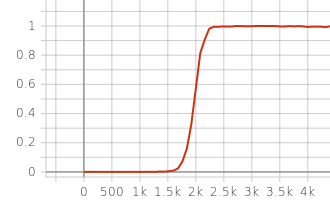}
    \end{subfigure}
    \begin{subfigure}[t]{0.45\textwidth}
        \includegraphics[width=\linewidth]{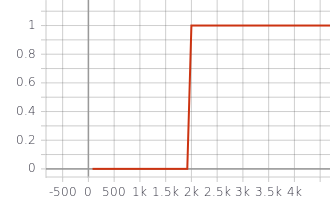}
    \end{subfigure}
    \captionsetup{skip=10pt}
    \caption{Example training (left) and evaluation (right) reward curves averaged over the batch dimension for an agent trained with PPO + ICM.}
    \label{fig:exp_1_plot}
\end{figure}

\begin{table}[t]
    \centering
    \begin{tabular}{ccc}
        \toprule
        Algorithm        & \# trials task is solved & $\mu \pm \sigma$ episodes to solve \\
        \midrule
        PPO              & 0 / 5 & - \\
        PPO + ICM        & 5 / 5 & 102K $\pm$ 60K \\
        \bottomrule
    \end{tabular}
    \captionsetup{skip=10pt}
    \caption{Number of trials in which the base task is solved, and mean ($\mu$) plus or minus one standard deviation ($\sigma$) episodes required to solve the base task, for PPO with and without ICM.}
    \label{tab:exp_1_results}
\end{table}

As described in \nameref{sec:exp_setup}, Experiment 1 consists of training an agent on the base environment, and quantifying its performance on all possible transfer sets. We train agents for 250 iterations, with a training batch size of 1024 episodes, or at most 12.8M transitions since episodes terminate early on successful completion of the task. We consider the task solved if the agent is able to solve it at least once during the evaluation phase, since its actions are deterministic and the task remains the same. Results are gathered in Table \ref{tab:exp_1_results}. The agent trained with PPO and ICM is able to solve the base task on all 5 trials, whereas, without ICM, the agent fails to solve it on any trial in the allocated time. This is likely due to the sparse nature of the reward in this environment, where intrinsic curiosity is needed to encourage exploration. Both agents fail to solve any of the transfer tasks except for when it is sampled such that it is identical to the base task (data not shown). Figure \ref{fig:exp_1_plot} depicts average training and evaluation rewards from an example training run for an agent training with PPO and ICM. Since evaluation is deterministic, the policy either completely succeeds or completely fails to solve the task.

\subsection{Experiment 2}

\begin{figure}[t]
    \centering
    \includegraphics[width=0.9\linewidth]{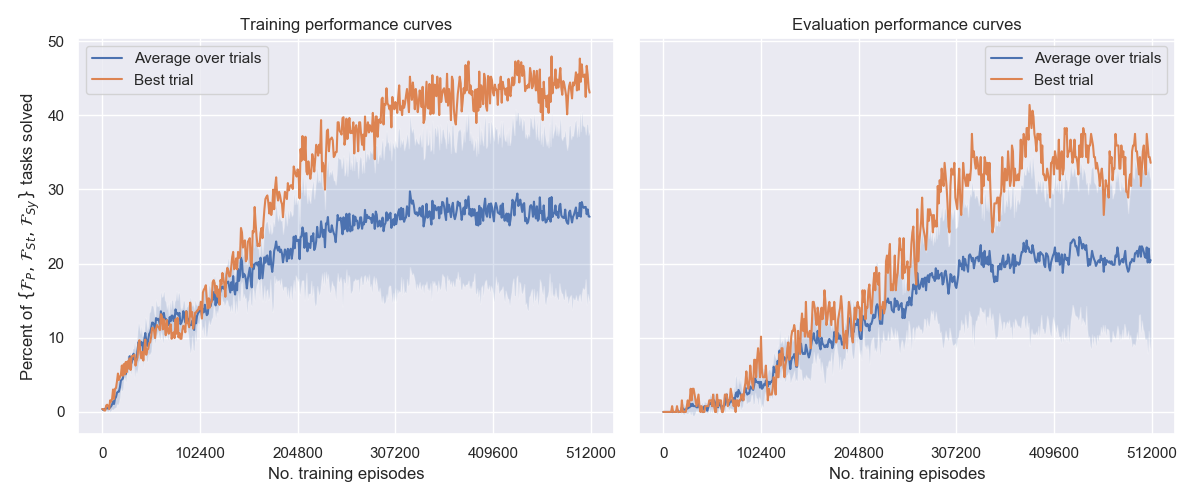}
    \caption{Mean $\pm$ 1 standard deviation and maximum training and evaluation performance curves from training 5 PPO + ICM agents to solve tasks from $\{\mathcal{F}_{P},\,\mathcal{F}_{St},\,\mathcal{F}_{Sy}\}$.}
    \label{fig:exp_2_per_curves}
\end{figure}

\begin{table}[t]
    \centering
    \resizebox{\columnwidth}{!}{
        \begin{tabular}{cccccccc}
            \toprule
            Algorithm        & $\{\mathcal{F}_{P}\}$ & $\{\mathcal{F}_{St}\}$ & $\{\mathcal{F}_{Sy}\}$ & $\{\mathcal{F}_{P},\,\mathcal{F}_{St}\}$ & $\{\mathcal{F}_{P},\,\mathcal{F}_{Sy}\}$ & $\{\mathcal{F}_{St},\,\mathcal{F}_{Sy}\}$ & $\{\mathcal{F}_{P},\,\mathcal{F}_{St},\,\mathcal{F}_{Sy}\}$ \\
            \midrule
            PPO              & 0\%                    & 0\% & 35.5\% $\pm$ 10.9\%          & 0\% & 7\% $\pm$ 3.6\%             & 19.4\% $\pm$ 3.2\%           & 4.4\% $\pm$ 3.2\%         \\
            \midrule
            PPO + ICM        & \textbf{2\% $\pm$ 3\%} & 0\% & \textbf{40.2\% $\pm$ 16.9\%} & 0\% & \textbf{29.7\% $\pm$ 8.1\%} & \textbf{33.1\% $\pm$ 19.7\%} & \textbf{24.4\% $\pm$ 9\%} \\
            \bottomrule
        \end{tabular}
    }
    \captionsetup{skip=10pt}
    \caption{Percentage of tasks solved (mean $\pm$ 1 standard deviation over 5 trials) during evaluation on each of the transfer sets from training on transfer set $\{\mathcal{F}_{P},\,\mathcal{F}_{St},\,\mathcal{F}_{Sy}\}$.}
    \label{tab:exp_2_results}
\end{table}

For Experiment 2, we train an agent on the transfer set $\{\mathcal{F}_{P},\,\mathcal{F}_{St},\,\mathcal{F}_{Sy}\}$. We use 500 training iterations, which amounts to at most 25.6M training transitions. Figure \ref{fig:exp_2_per_curves} plots the percentage of training and evaluation tasks solved over the course of training for agents trained with PPO + ICM. Table \ref{tab:exp_2_results} reports the average percentage of solved tasks over all transfer sets for both PPO and PPO + ICM agents at the point in training where the highest overall evaluation is obtained. The PPO + ICM agent outperforms the plain PPO agent, again possibly due to sparse rewards in all transfer environments.

The failure of either agent to completely solve any single transfer set (exhibiting a particular kind of reasoning) demonstrates the difficulty of the environments for existing RL methods. We expect that, given enough compute, these environments may be conquered by the agents described above without modification. However, we hope that this work stimulates research into priors that encourage different modes of reasoning that result in improved sample efficiency and structured exploration.

\section{Discussion}

In this section, we discuss the gap of research and ideas that exist between the fields of RL and animal tool use. Although it is not yet evident that there exist methods that provide the prior reasoning capabilities needed to learn tool using behaviors in general, we provide some insight into why this may be and the mechanisms that might be required.

\subsection{Bridging the gap between RL and tool use}

Animals and humans alike use tools to solve otherwise unsolvable problems, or to arrive at more efficient or safe solutions that are out of reach without certain tools \cite{Seed2010}. In the same way, we conjecture that RL agents could benefit by leveraging tools, for example by extending their reach with a stick object as in the simulated trap-tube task. Tools effectively modify the effects of an agent's actions, and accordingly, when equipped, it affects the degree to which it can control its environment. There is evidence in humans and monkeys that using tools as an extension of the body is followed by changes in neural networks that represent body schema \cite{Maravita2004}, likely due to the change in mechanical and sensory capabilities they afford. Tools that enable better control over the environment should allow RL agents to discover solutions to problems that take fewer steps, or to solve problems which, without them, cannot be solved at all.

How RL agents make use of objects in their environment can be characterized in the language of tool use: What objects can be considered tools? Does the agent employ relational, causal, or other modes of reasoning to use tools to make tasks easier or solve-able? An agent's behavior can be studied from the perspective of tool use researchers, and can be updated to better match the learning or exploration behavior exemplified by model organisms. In this way, tool use researchers may build simulation models of the real-world phenomena that they wish to study, enabling rapid iteration and intuition building on the mechanisms behind tool use.

We claim that many of the tasks studied in the tool use literature may be cast in the RL framework, and indeed should be in order to test hypotheses concerning the cognitive processes underlying tool use. This work takes a step towards realizing this goal by implementing a simulation of the well-studied trap-tube task. As mentioned above, the variations on the trap-tube task are well-suited to studying different forms of generalization, all which have real-world counterparts in human and non-human animal tool use experiments.

\subsection{Potential prior reasoning capabilities}

Each type of generalization introduced in this paper requires specific characteristics of agent behavior. At present, to develop these characteristics, agents are only sparsely and extrinsically incentivized to learn to solve the tasks presented. We ask the question: \textit{What priors are required to solve tool use in general?} The RL framework depends heavily on the reward signal provided to the agent over time. In the case studied in this paper, the reward signal is sparse and is only seen by the agent upon successfully completing the task. Without strong behavioral and reasoning priors, agents will arrive at tool use seemingly at random and are unreasonably expected to generalize from limited exploratory experience.

For perceptual reasoning abilities, using a convolutional neural network (CNN) may be a strong prior. A CNN fits the modality presented to the agent (i.e., observations of a grid world) and the physically local nature of objects, but alone may not enable learning of tool using behaviors due to the non-local nature of object-object interactions.

A structural reasoning prior may not be as easily understood due to the nature of the feature space provided. In the structural task, agents must \textit{remember} the representation of objects in order to reason about a solution. Take for example the structural representations of the \texttt{trap} and the \texttt{exit}: since are always parallel to each other and adjacent to the \texttt{tube}, there may be no way to determine, without interaction, which object the \texttt{food} will pass through. Thus, the agent must \textit{remember} which object-object interactions it has explored during each new episode, possibly developing a causal understanding of the environment.

Symbolic reasoning with tools is much less understood than perceptual and structural reasoning, but may require similar priors. Agents must learn to discover the identity and proper utilization of each object and solve the task through changing object-object interactions on each new episode.

\section{Conclusion}

In this paper, we reviewed aspects of tool use from the perspective of several fields of study and ultimately chose to frame it as a RL problem. With the support of a classic trap-tube experiment as an example, we presented a framework for analyzing the generalization of RL agents, taking a step towards understanding the priors required to learn to use tools. Furthermore, we defined two experimental setups, allowing researchers to define their own hypotheses for understanding tool use in the form of a RL generalization problem. Finally, we provided baselines using the well-established proximal policy optimization algorithm to compare against in future work.

We chose to focus on the definition of \textit{tool use}, omitting many topics related to tool use such as \textit{proto-tool use}, \textit{meta-tool use} and \textit{tool manufacture}. We briefly outline these for the reader (see \nameref{section:appendicies_definitions}) and leave exploration of the differences between them as they relate to RL for future work.

\subsection*{Acknowledgements}

We gratefully thank Danielle Swank for her dedicated mediation in the discussion of the research process behind this paper. We also thank Ed Costantini and Andrew Wang for their careful review of and helpful comments on the paper.

\bibliography{references}

\begin{thebibliography}{10}

\bibitem{Tebbich2010}
Sabine Tebbich, Kim Sterelny, and Irmgard Teschke.
\newblock The tale of the finch: adaptive radiation and behavioural
  flexibility.
\newblock {\em Philosophical Transactions of the Royal Society B: Biological
  Sciences}, 365(1543):1099--1109, April 2010.

\bibitem{Krutzen2005}
M.~Krutzen, J.~Mann, M.~R. Heithaus, R.~C. Connor, L.~Bejder, and W.~B.
  Sherwin.
\newblock Cultural transmission of tool use in bottlenose dolphins.
\newblock {\em Proceedings of the National Academy of Sciences},
  102(25):8939--8943, June 2005.

\bibitem{Breuer2005}
Thomas Breuer, Mireille Ndoundou-Hockemba, and Vicki Fishlock.
\newblock First observation of tool use in wild gorillas.
\newblock {\em {PLoS} Biology}, 3(11):e380, October 2005.

\bibitem{Teschke2013}
I.~Teschke, C.~A.~F. Wascher, M.~F. Scriba, A.~M.~P. von Bayern, V.~Huml,
  B.~Siemers, and S.~Tebbich.
\newblock Did tool-use evolve with enhanced physical cognitive abilities?
\newblock {\em Philosophical Transactions of the Royal Society B: Biological
  Sciences}, 368(1630):20120418--20120418, October 2013.

\bibitem{Visalberghi1994}
Elisabetta Visalberghi and Luca Limongelli.
\newblock Lack of comprehension of cause-effect relations in tool-using
  capuchin monkeys (cebus apella).
\newblock {\em Journal of Comparative Psychology}, 108(1):15--22, 1994.

\bibitem{McCormack2011}
Teresa McCormack, Christoph Hoerl, and Stephen Butterfill, editors.
\newblock {\em Tool Use and Causal Cognition}.
\newblock Oxford University Press, August 2011.

\bibitem{Reaux2003}
James~E. Reaux and Daniel~J. Povinelli.
\newblock The trap-tube problem.
\newblock In {\em Folk Physics for Apes}, pages 108--131. Oxford University
  Press, May 2003.

\bibitem{Povinelli2011}
Daniel~J. Povinelli and Derek~C. Penn.
\newblock Through a floppy tool darkly.
\newblock In {\em Tool Use and Causal Cognition}, pages 69--88. Oxford
  University Press, August 2011.

\bibitem{Penn2008}
Derek~C. Penn, Keith~J. Holyoak, and Daniel~J. Povinelli.
\newblock Darwin's mistake: Explaining the discontinuity between human and
  nonhuman minds.
\newblock {\em Behavioral and Brain Sciences}, 31(2):109--130, April 2008.

\bibitem{DeLoache1997}
Judy~S DeLoache, Kevin~F. Miller, and Karl~S. Rosengren.
\newblock The credible shrinking room: Very young children's performance with
  symbolic and nonsymbolic relations.
\newblock {\em Psychological Science}, 8(4):308--313, July 1997.

\bibitem{Seed2010}
Amanda Seed and Richard Byrne.
\newblock Animal tool-use.
\newblock {\em Current Biology}, 20(23):R1032--R1039, dec 2010.

\bibitem{Li1988}
Z.~Li and S.S. Sastry.
\newblock Task-oriented optimal grasping by multifingered robot hands.
\newblock {\em {IEEE} Journal on Robotics and Automation}, 4(1):32--44, 1988.

\bibitem{Shimoga1996}
K.B. Shimoga.
\newblock Robot grasp synthesis algorithms: A survey.
\newblock {\em The International Journal of Robotics Research}, 15(3):230--266,
  June 1996.

\bibitem{Yamashita1998}
Atsushi Yamashita, Jun Sasaki, Jun Ota, and Tamio Arai.
\newblock Cooperative manipulation of objects by multiple mobile robots with
  tools *.
\newblock 1998.

\bibitem{Gupta1998}
S.K. Gupta, C.J.J. Paredis, and P.F. Brown.
\newblock Micro planning for mechanical assembly operations.
\newblock In {\em Proceedings. 1998 {IEEE} ICRA (Cat. No.98CH36146)}. {IEEE}.

\bibitem{Halperin2000}
D.~Halperin, J.-C. Latombe, and R.~H. Wilson.
\newblock A general framework for assembly planning: The motion space approach.
\newblock {\em Algorithmica}, 26(3-4):577--601, March 2000.

\bibitem{Stoytchev2005}
A.~Stoytchev.
\newblock Behavior-grounded representation of tool affordances.
\newblock In {\em Proceedings of the 2005 {IEEE} ICRA}. {IEEE}.

\bibitem{Brown2012}
Solly Brown and Claude Sammut.
\newblock Tool use and learning in robots.
\newblock In {\em Encyclopedia of the Sciences of Learning}, pages 3327--3330.
  Springer {US}, 2012.

\bibitem{Wicaksono2016}
Handy Wicaksono and Claude Sammut.
\newblock Relational tool use learning by a robot in a real and simulated
  world.
\newblock 2016.

\bibitem{Wicaksono2017}
Handy Wicaksono.
\newblock Towards a relational approach for tool creation by robots.
\newblock In {\em Proceedings of the Twenty-Sixth International Joint
  Conference on Artificial Intelligence}. International Joint Conferences on
  Artificial Intelligence Organization, August 2017.

\bibitem{Lenz2015}
Ian Lenz, Ross~A. Knepper, and Ashutosh Saxena.
\newblock Deepmpc: Learning deep latent features for model predictive control.
\newblock In {\em Robotics: Science and Systems}, 2015.

\bibitem{Fang2018}
Kuan Fang, Yuke Zhu, Animesh Garg, Andrey Kurenkov, Viraj Mehta, Li~Fei{-}Fei,
  and Silvio Savarese.
\newblock Learning task-oriented grasping for tool manipulation from simulated
  self-supervision.
\newblock {\em CoRR}, abs/1806.09266, 2018.

\bibitem{Rajeswaran2017}
Aravind Rajeswaran, Vikash Kumar, Abhishek Gupta, John Schulman, Emanuel
  Todorov, and Sergey Levine.
\newblock Learning complex dexterous manipulation with deep reinforcement
  learning and demonstrations.
\newblock {\em CoRR}, abs/1709.10087, 2017.

\bibitem{Ebert2017}
Frederik Ebert, Chelsea Finn, Alex~X. Lee, and Sergey Levine.
\newblock Self-supervised visual planning with temporal skip connections.
\newblock {\em CoRR}, abs/1710.05268, 2017.

\bibitem{Ebert2018}
Frederik Ebert, Chelsea Finn, Sudeep Dasari, Annie Xie, Alex~X. Lee, and Sergey
  Levine.
\newblock Visual foresight: Model-based deep reinforcement learning for
  vision-based robotic control.
\newblock {\em CoRR}, abs/1812.00568, 2018.

\bibitem{Xie2019}
Annie Xie, Frederik Ebert, Sergey Levine, and Chelsea Finn.
\newblock Improvisation through physical understanding: Using novel objects as
  tools with visual foresight.
\newblock {\em CoRR}, abs/1904.05538, 2019.

\bibitem{Sutton2018}
Richard~S. Sutton and Andrew~G. Barto.
\newblock {\em Reinforcement Learning: An Introduction}.
\newblock The MIT Press, second edition, 2018.

\bibitem{Farebrother2018}
Jesse Farebrother, Marlos~C. Machado, and Michael Bowling.
\newblock Generalization and regularization in {DQN}.
\newblock {\em CoRR}, abs/1810.00123, 2018.

\bibitem{Packer2018}
Charles Packer, Katelyn Gao, Jernej Kos, Philipp Kr{\"{a}}henb{\"{u}}hl,
  Vladlen Koltun, and Dawn Song.
\newblock Assessing generalization in deep reinforcement learning.
\newblock {\em CoRR}, abs/1810.12282, 2018.

\bibitem{Cobbe2018}
Karl Cobbe, Oleg Klimov, Christopher Hesse, Taehoon Kim, and John Schulman.
\newblock Quantifying generalization in reinforcement learning.
\newblock {\em CoRR}, abs/1812.02341, 2018.

\bibitem{StAmant2008}
Robert~St Amant and Thomas~E. Horton.
\newblock Revisiting the definition of animal tool use.
\newblock {\em Animal Behaviour}, 75(4):1199--1208, apr 2008.

\bibitem{Kacelnik2009b}
Alex Kacelnik, Jackie Chappell, Ben Kenward, and Alex A.~S. Weir.
\newblock Cognitive adaptations for tool-related behavior in new caledonian
  crows.
\newblock In {\em Comparative {CognitionExperimental} Explorations of Animal
  Intelligence}, pages 515--528. Oxford University Press, April 2009.

\bibitem{Seed2009}
Amanda~M. Seed, Josep Call, Nathan~J. Emery, and Nicola~S. Clayton.
\newblock Chimpanzees solve the trap problem when the confound of tool-use is
  removed.
\newblock {\em Journal of Experimental Psychology: Animal Behavior Processes},
  35(1):23--34, 2009.

\bibitem{Seed2011}
Amanda Seed, Daniel Hanus, and Josep Call.
\newblock Causal knowledge in corvids, primates, and children.
\newblock In {\em Tool Use and Causal Cognition}, pages 89--110. Oxford
  University Press, August 2011.

\bibitem{Schulman2017}
John Schulman, Filip Wolski, Prafulla Dhariwal, Alec Radford, and Oleg Klimov.
\newblock Proximal policy optimization algorithms.
\newblock {\em CoRR}, abs/1707.06347, 2017.

\bibitem{Pathak2017}
Deepak Pathak, Pulkit Agrawal, Alexei~A. Efros, and Trevor Darrell.
\newblock Curiosity-driven exploration by self-supervised prediction.
\newblock {\em CoRR}, abs/1705.05363, 2017.

\bibitem{Cho2014}
KyungHyun Cho, Bart van Merrienboer, Dzmitry Bahdanau, and Yoshua Bengio.
\newblock On the properties of neural machine translation: Encoder-decoder
  approaches.
\newblock {\em CoRR}, abs/1409.1259, 2014.

\bibitem{Maravita2004}
Angelo Maravita and Atsushi Iriki.
\newblock Tools for the body (schema).
\newblock {\em Trends in Cognitive Sciences}, 8(2):79 -- 86, 2004.

\bibitem{LawickGoodall1971}
Hugo~Van Lawick and Jane Goodall.
\newblock {\em Innocent Killers}.
\newblock Houghton Mifflin, 1971.

\bibitem{Alcok1972}
John Alcock.
\newblock The evolution of the use of tools by feeding animals.
\newblock {\em Evolution}, 26(3):464--473, 1972.

\bibitem{Beck1980}
Benjamin~B. Beck.
\newblock {\em Animal Tool Behavior: The Use and Manufacture of Tools by
  Animals}.
\newblock Garland STPM Press, 1980.

\bibitem{Taylor2007}
Alex~H. Taylor, Gavin~R. Hunt, Jennifer~C. Holzhaider, and Russell~D. Gray.
\newblock Spontaneous metatool use by new caledonian crows.
\newblock {\em Current Biology}, 17(17):1504--1507, September 2007.

\end{thebibliography}

\newpage
\appendix

\section{Appendix: Definitions} \label{section:appendicies_definitions}

\subsection{Tool Use}

Although there are many proposed tool use definitions, in this paper we have decided that the Amant and Horton \cite{StAmant2008} definition is most representative of our study. There are other definitions that have inspired this work; we use the first as the working definition for this paper:

\begin{displayquote}
The exertion of control over a freely manipulable external object (the tool) with the goal of (1) altering the physical properties of another object, substance, surface or medium (the target, which may be the tool user or another organism) via a dynamic mechanical interaction, or (2) mediating the flow of information between the tool user and the environment or other organisms in the environment. - Amant and Horton \cite{StAmant2008}
\end{displayquote}

\begin{displayquote}
The use of an external object as a functional extension of mouth or beak, hand or claw, in the attainment of an immediate goal. — Lawick and Goodall \cite{LawickGoodall1971}, page 195
\end{displayquote}

\begin{displayquote}
Tool-using involves the manipulation of an inanimate object, not internally manufactured, with the effect of improving the animal’s efficiency in altering the form or position of some separate object. — Alcock \cite{Alcok1972}, page 464
\end{displayquote}

\begin{displayquote}
Thus tool use is the external employment of an unattached environmental object to alter more efficiently the form, position, or condition of another object, another organism, or the user itself when the user holds or carries the tool during or just prior to use and is responsible for the proper and effective orientation of the tool. — Beck \cite{Beck1980}, page 10
\end{displayquote}

\subsection{Proto-Tool Use}

\begin{displayquote}
Proto-tool use is distinguished from "true" tool use in that the outcome is achieved via a secondary object or substance, albeit not something defined as a tool. — Amant and Horton \cite{StAmant2008}
\end{displayquote}

\noindent{The distinction between proto-tool use and "true" tool use is made in behavioral neurophysiology because animals exhibiting the proto-tool use tend to have smaller brains than "true" tool users.}

\subsection{Meta-Tool Use}

\begin{displayquote}
The ability to use one tool on another. – Taylor et al. \cite{Taylor2007}
\end{displayquote}

\noindent{For our purposes, we define meta-tool use as the utilization of a static object in a preparatory behavior for tool use.}

\subsection{Tool Manufacture}

\begin{displayquote}
Tool manufacture involves the fashioning or modification of objects in the environment to improve their suitability as tools. — Amant and Horton \cite{StAmant2008}
\end{displayquote}

\noindent{Tool manufacture is essentially multi-step tool use involving a preparation phase where the tool is manufactured and an application phase where the tool is applied.}

\end{document}